\documentclass[11pt, a4paper, logo, twocolumn, copyright]{googledeepmind}

\usepackage[authoryear, sort&compress, round]{natbib}
\providecommand{\wt}{1h-walk VQA}

\title{Perception Test 2024: Challenge Summary and a Novel Hour-Long VideoQA Benchmark}

\correspondingauthor{viorica@google.com}


\author[1]{Joseph Heyward}
\author[1]{Jo\~ao Carreira}
\author[1,2]{Dima Damen}
\author[1,3]{Andrew Zisserman}
\author[1]{Viorica P\u atr\u aucean}

\affil[1]{Google DeepMind}
\affil[2]{University of Bristol}
\affil[3]{University of Oxford}

\begin{abstract}
Following the successful 2023 edition, we organised the Second Perception Test challenge as a half-day workshop alongside the IEEE/CVF European Conference on Computer Vision (ECCV) 2024, with the goal of benchmarking state-of-the-art video models and measuring the progress since last year using the Perception Test benchmark.
This year, the challenge had seven tracks (up from six last year) and covered low-level and high-level tasks, with language and non-language interfaces, across video, audio, and text modalities; the additional track covered hour-long video understanding and introduced a novel video QA benchmark \emph{\wt}. Overall, the tasks in the different tracks were: object tracking, point tracking, temporal action localisation, temporal sound localisation, multiple-choice video question-answering, grounded video question-answering, and hour-long video question-answering. We summarise in this report the challenge tasks and results, and introduce in detail the novel hour-long video QA benchmark \emph{\wt}.
\end{abstract}

\begin{document}

\maketitle

\section{Introduction}
\begin{figure}
    \centering
    \includegraphics[width=\linewidth]{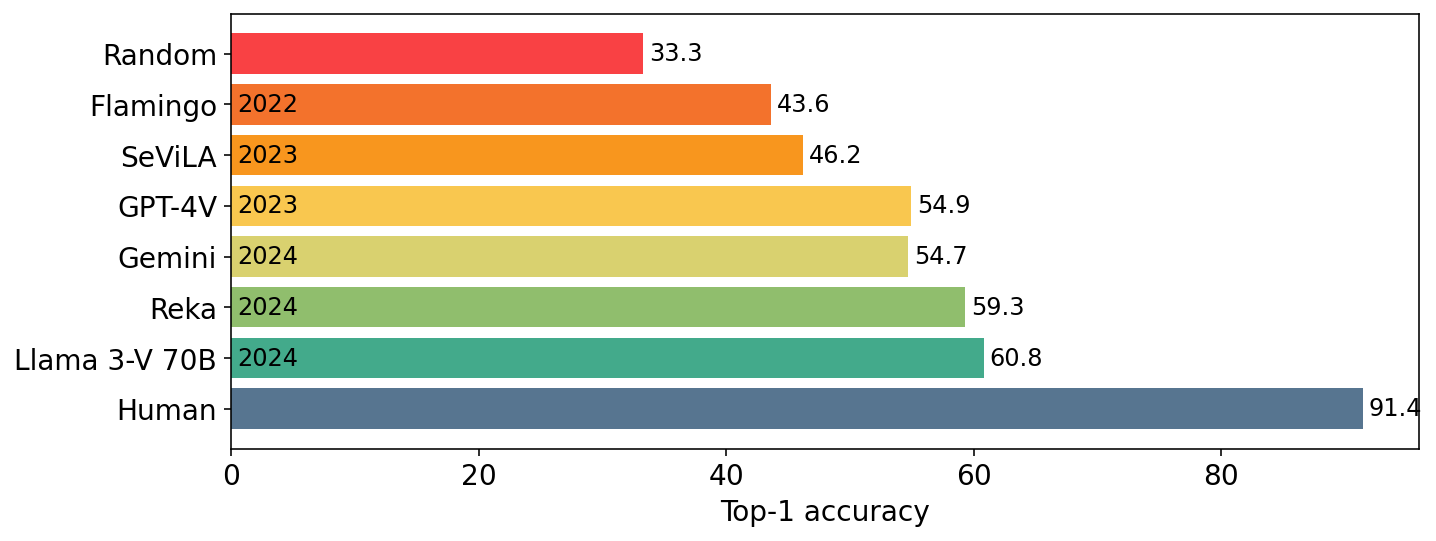}
    \caption{Top-1 accuracy of recent VLMs vs human baseline on the Perception Test multiple-choice video QA task. We include the results published by models' authors where available, otherwise we ran the models independently (GPT-4V, SeViLA, Flamingo).}
    \label{fig:sotahuman}
\end{figure}

Multimodal video models have witnessed a tremendous boost in performance these past couple of years, with both proprietary and open-sourced models pushing the boundaries of machine perception capabilities, e.g., Flamingo~\citep{alayrac2022flamingo}, SeViLA~\citep{yu2023self}, GPT-4V~\citep{2023GPT4VisionSC}, Gemini~\citep{geminiteam2024geminifamilyhighlycapable}, Reka~\citep{rekateam2024rekacoreflashedge}, Llama 3-V~\citep{dubey2024llama3herdmodels}. In 2023, we introduced the Perception Test benchmark~\citep{patraucean2023perception} to comprehensively measure the performance of video models on different perception tasks and across modalities. It can be observed that the performance of video-language models is steadily increasing over time on the video-language tracks in our benchmark, but there is still a significant gap compared to human performance; see Figure~\ref{fig:sotahuman}.  
Additionally, other tasks such as tracking and temporal segmentation still require specialised models with handcrafted pipelines.
To keep track of progress over time, we set up a yearly public challenge using our benchmark and we invite participants to submit their best model's predictions. This year, we organised the second edition as a workshop at ECCV 2024, featuring 7 challenge tracks (compared to six tracks at the first edition).  

\noindent \textbf{Benchmark:} 
The \textit{Perception Test}~\citep{patraucean2023perception} is a comprehensive benchmark that uses purposefully-designed real-world videos to diagnose perception capabilities like memory, understanding of intuitive physics and geometry, abstract patterns, and semantics. The benchmark consists of 11.6k videos, with audio, up to 35s long, filmed by diverse crowd-sourced participants following scripts designed to show perceptually-interesting situations. The focus is on probing generalisation and transfer capabilities, so the benchmark only provides a relatively small training set to be used for fine-tuning or prompting, and the rest is used for evaluation. The videos have six types of annotations enabling language and non-language evaluations, across video, audio, and text modalities. More details about the Perception Test and data samples are available on our github repository\footnote{\url{https://github.com/google-deepmind/perception_test}} and on the workshop website\footnote{\url{https://ptchallenge-workshop.github.io/}}.

\noindent \textbf{Additional benchmark:} In addition, this year, to assess models' capability of reasoning over very long temporal context, we introduce \emph{\wt} -- a novel small-scale benchmark based on the Walking Tours dataset~\citep{venkataramanan2023imagenet}; see details in Section~\ref{sec:wt}.  

\noindent \textbf{Challenge tracks:} The videos in the Perception Test benchmark are annotated with the following human-collected labels: object tracks, point tracks, action segments, sound segments, multiple-choice video question-answers, and grounded video question-answers; the additional dataset included this year was annotated with multiple-choice video question-answers. For each type of annotation, we define a corresponding challenge track. We describe in the next sections the setup, metrics, and results in each track.    

\noindent \textbf{Challenge setup:} We relied on the open-source eval.ai platform to set up the different challenge tracks. Each track had 2 phases (validation and test), each phase using the corresponding validation and test splits of the Perception Test benchmark and the newly added dataset. For each submission, the participants had to indicate the evaluation mode (fine-tuning, few-shot, or zero-shot evaluation). In some tracks, the participants had to indicate if the model used the audio modality as well or not (for action and sound localisation, multiple-choice video QA). For test submissions, the participants were required to also upload a short report describing their method (architecture, pre-training datasets and tasks, etc.).
The validation phase served as a sanity check for participants' submission pipelines. The number of submissions for the validation phase was not limited.

The test set was made available 2.5 months before the submission deadline. 
For the test phase, the limit was set to 2 submissions per day, 30 submissions in total. 
Only the results made public on the test leaderboard were considered for the competition. 

\section{\textit{\wt}: A Novel Hour-Long VideoQA Benchmark}
\label{sec:wt}

\begin{figure}
    \centering
    \includegraphics[width=\linewidth]{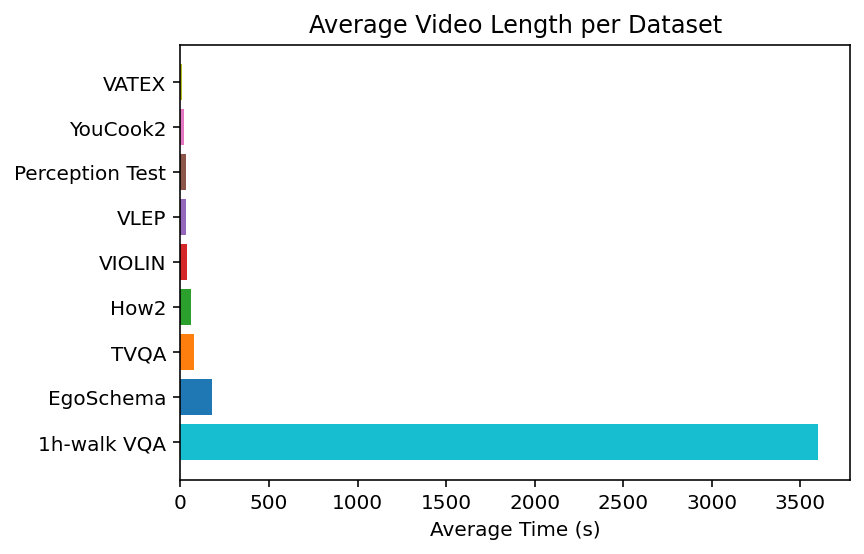}
    \caption{Average video length in our newly-proposed \textit{\wt}\ benchmark compared to existing benchmarks.}
    \label{fig:hourlengths}
\end{figure}

\begin{figure*}
    \centering
    \includegraphics[width=\linewidth]{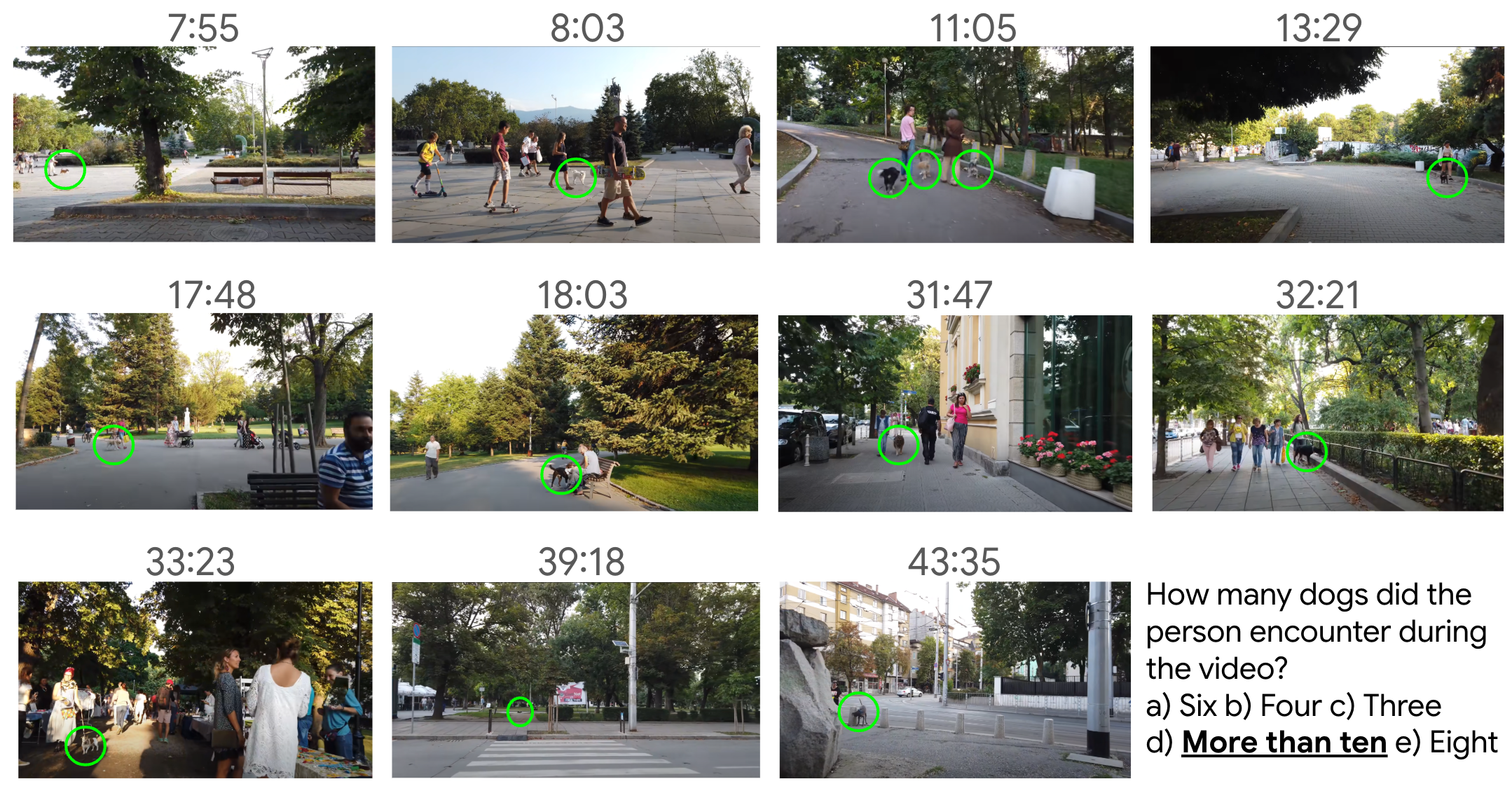}
    \caption{Example of a counting question in \textit{\wt}\ that spans more than 30 minutes. We show the relevant frames and their associated timestamps. The correct answer is marked in \underline{\textbf{bold}}.}
    \label{fig:dogs}
\end{figure*}

We rely on the Walking Tours dataset~\citep{venkataramanan2023imagenet} to create a small-scale but very challenging benchmark to assess models' ability to understand and reason over very long temporal contexts (hour-long).
The Walking Tours dataset contains ten 1-hour (or longer) Youtube videos with natural audio (no narrations\footnote{It is important that these videos are not narrated to ensure no shortcut through language can be used.}), that depict city tours filmed by people while walking around different cities. Figure~\ref{fig:hourlengths} shows a comparison in terms of video length between the proposed benchmark and existing datasets. We augment this dataset with 70 manually-curated challenging 5-way question-answer pairs that require reasoning over video and/or audio modalities. We name \emph{\wt}\ the resulting benchmark. 

Collecting challenging questions that span long temporal contexts is very difficult, even for humans. Often, the questions in existing benchmarks can be answered from a single frame or a short clip~\citep{Papalampidi2023ASR}. To ensure that our questions require long context, we ran several iterations of annotation collection with human raters. In a first iteration, each rater was tasked to watch an hour-long video and propose different types of questions: 2 questions that require one video segment to be answered, 2 questions that require 2 temporally-separated video segments to be answered, 1 question that requires more than 2 video segments to be answered, and 1 question that requires video and audio to be answered. Our team manually reviewed all the provided questions and selected those that cannot be answered from a single frame or a very short clip. We then ran a second iteration of annotations, more targeted to particular events, where we first ran a detection step to localise in time particular (repeated) events and then we designed questions based on those timestamps. For example, we asked raters to mark all the video segments where the person wearing the camera crosses a bridge, or walks up some stairs; or when a tower clock is visible in the video, or a distinct sound can be heard. We include in the appendix the list of unique questions selected for our final \emph{\wt}\ benchmark and we provide in Figure~\ref{fig:dogs} an example of a counting question that spans more than 30 minutes. More visualisations can be found on the challenge website\footnote{\url{https://eval.ai/web/challenges/challenge-page/2330/overview}}. 

This small benchmark is intended for zero-shot evaluation. We do not provide any training or fine-tuning data. We only provide a very small validation split to be used for sanity checks in the public challenge; see Table~\ref{tab:hvqa}. 
\begin{table}[]
    \centering
    \begin{tabular}{l|r|r}
      \textbf{Split} & \textbf{\# videos} & \textbf{\# questions} \\
       \hline
      Train & -  & - \\
      Validation  & 3 & 11 \\
      Test  & 7 & 59 \\
      \hline 
    \end{tabular}
    \caption{Splits in \textit{\wt}\ benchmark used for the hour-long video QA task.}
    \label{tab:hvqa}
\end{table}

\section{Overall Results Summary}  
\begin{figure}
    \centering
    \includegraphics[width=\linewidth]{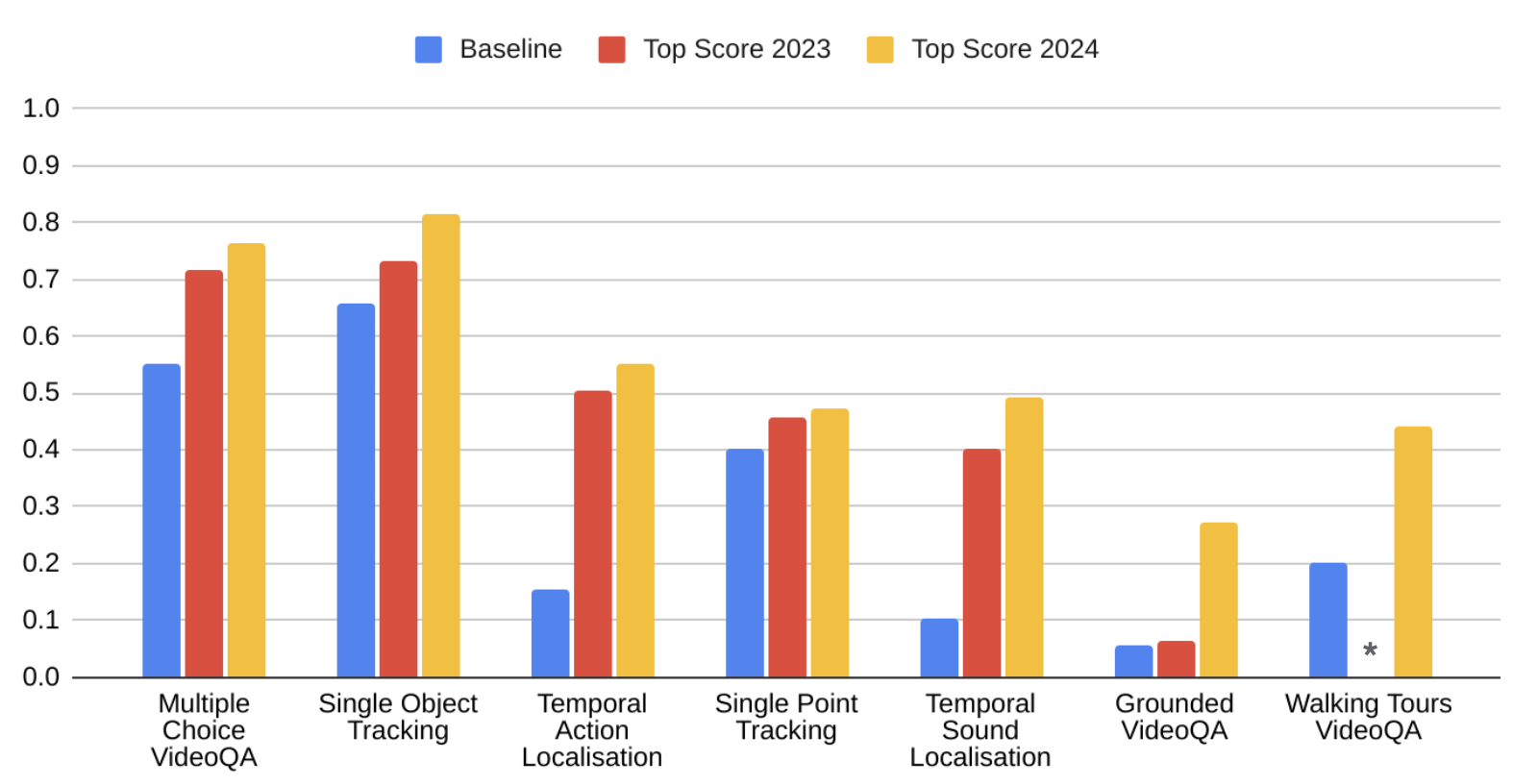}
    \caption{Per-track performance improvement compared to baselines and compared to best models from 2023, respectively.}
    \label{fig:summary2024}
\end{figure}

\begin{figure}
    \centering
    \includegraphics[width=\linewidth]{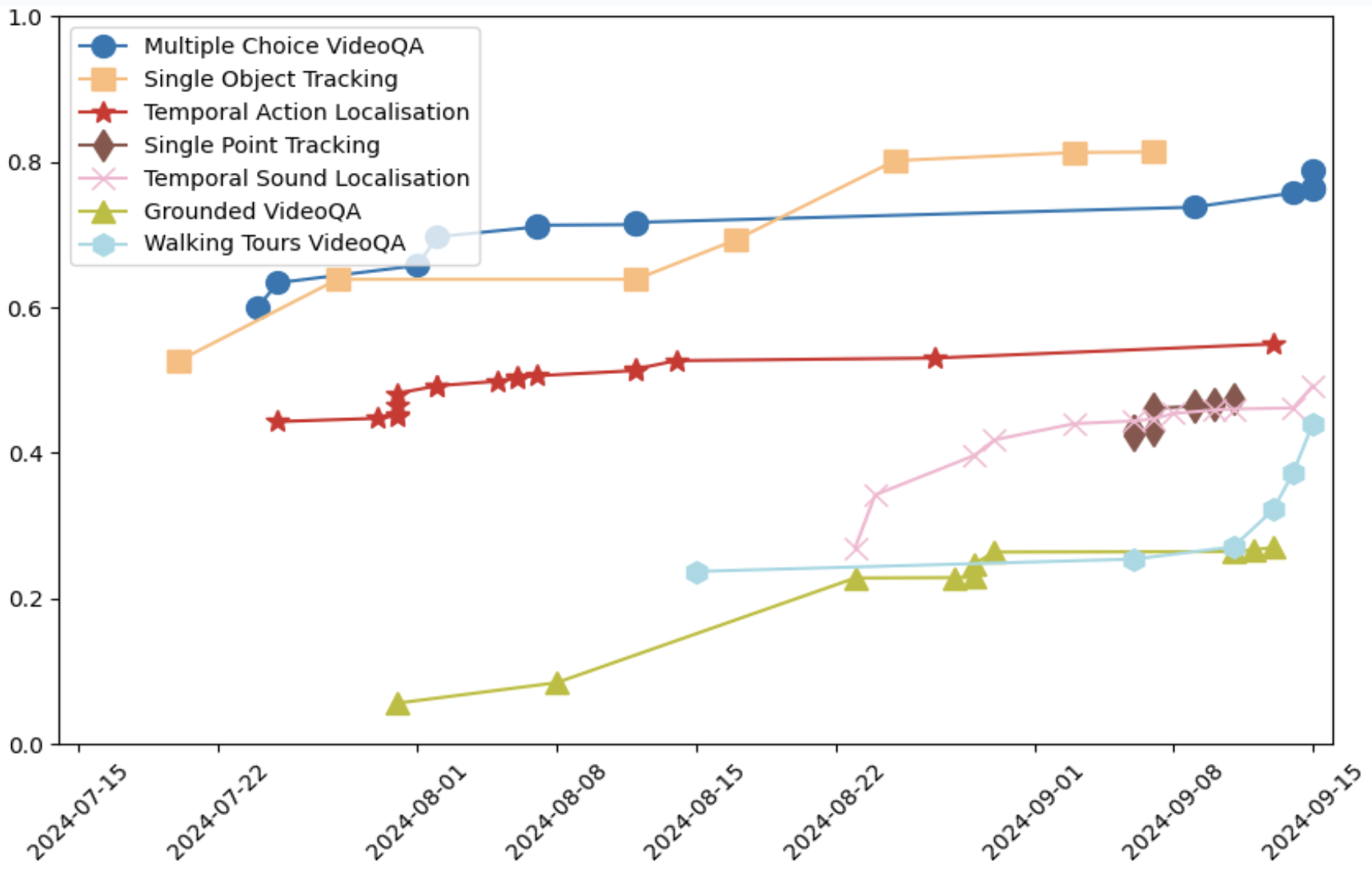}
    \caption{Per-task performance improvement of top models during the 2024 test submission phase.}
    \label{fig:pertask}
\end{figure}

We received 680 submissions from 123 teams across all seven tracks in both phases, up from 475 submissions from 63 teams in 2023. We awarded 2 prizes per track (best and runner-up) to submissions that obtained the best (and second best) results in the test leaderboard, with prizes totalling 20k EUR (up from 15k EUR in 2023). The top performing models improved when compared to the winning models from last year in all tracks. Figure~\ref{fig:summary2024}. Figure~\ref{fig:pertask} shows the evolution of the top-performing models during the test submission phase of this year's edition for each track. The reports of the winning submissions are available on the workshop website.

\section{Challenge Tracks, Results, Awards}
In the following we describe each track and the performance achieved in the challenge. For the technical report per team, including winners' affiliations and names, please refer to the workshop website: \url{https://ptchallenge-workshop.github.io/}.

\subsection{Object tracking}

\noindent \textbf{Task description:} For this task, the model receives a video and a bounding box representing an object, and it is required to track the object throughout the video sequence.

\noindent \textbf{Metric:} The evaluation metric for this task is average Intersection over Union (IoU). It is calculated as the average intersection over union between the predicted bounding boxes and the ground truth bounding boxes for each tracked object.

\noindent \textbf{Dataset:} As in the 2023 edition, to make the evaluation task more accessible, we used only a randomly selected subset of 1000 videos from the validation split of the Perception Test for the validation phase, and 1000 videos from the test split of the Perception Test for the test phase. We kept the same selection of videos as in the 2023 edition.

\noindent \textbf{Baselines:} We provide a simple dummy baseline for this task, which always assumes that the object is static, i.e. it outputs as predictions the initial bounding box received as input.

\begin{figure}
    \centering
    \includegraphics[width=\linewidth]{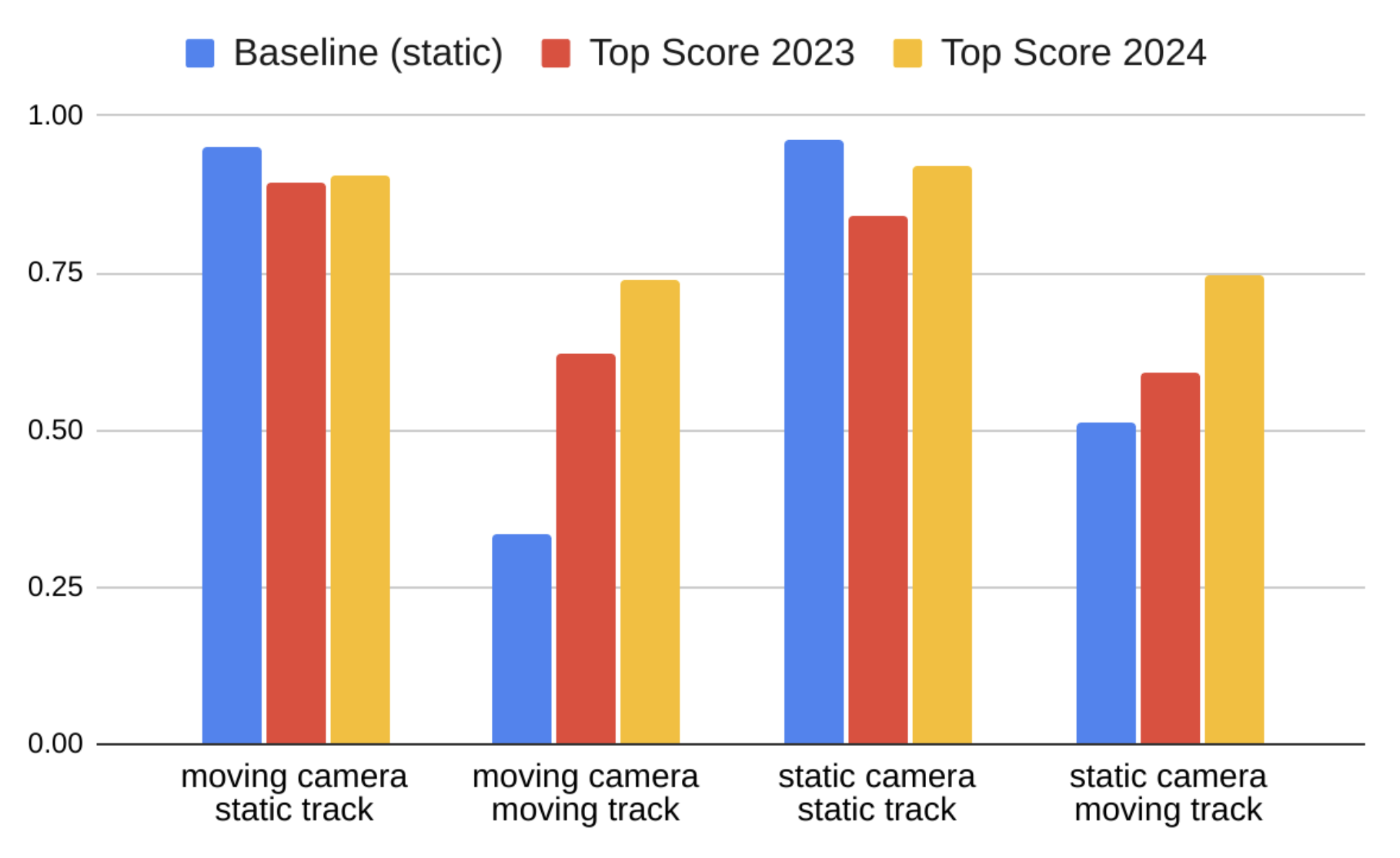}
    \caption{Baseline vs best results 2023 vs best results 2024 split by camera and object motion for the object tracking task.}
    \label{fig:sot}
\end{figure}

\noindent \textbf{Results:}
The results for the top-2 competing models are compared to the baseline in Table~\ref{tab:obj_tracking}. The top performing model relies on the recent LORAT~\citep{lorat} and shows a good improvement over the best submission from last year on both moving objects and moving camera categories in our dataset; see Figure~\ref{fig:sot} and check the authors' report on our workshop page for more details.  

\begin{table}[]
    \centering
    \begin{tabular}{l|l|c}
       \textbf{Rank} & \textbf{Team name} & \textbf{IoU} \\
       \hline
       Baseline & Dummy static & 0.640\\
       Runner-up & FAUgeddaboudit & 0.813 \\
       Best & NJUST-THU & 0.734 \\
         \hline
    \end{tabular}
    \caption{Object tracking results}
    \label{tab:obj_tracking}
\end{table}

\subsection{Point tracking}

\noindent \textbf{Task description:} In the single point tracking task, the model receives a video and the 2D coordinates of a point, and it is required to track the point throughout the video sequence, also accounting for occlusions.

\noindent \textbf{Metric:} The evaluation metric for this challenge is the average Jaccard, proposed in TAP-Vid~\citep{doersch2022tapvid}. It takes into account the Occlusion Accuracy -- a simple classification accuracy for the point occlusion prediction on each frame, and the Position accuracy -- for frames where the point is visible, it measures the fraction of points that are within a certain threshold of their ground truth; it assumes that the images are resized to 256x256 pixels and the accuracy is averaged across 5 thresholds: 1, 2, 4, 8, and 16 pixels. The final Jaccard metric calculates the fraction of \textit{true positives}, which are points within the threshold of any visible ground truth points, divided by \textit{true positives} plus \textit{false positives} (points that are predicted as visible but the ground truth is either occluded or farther than the threshold) plus \textit{false negatives} (ground truth visible points that are predicted as occluded or the prediction is farther than the threshold). The overall metric is Jaccard averaged across all thresholds.

\noindent \textbf{Dataset:} We use the same dataset as in 2023 for this task, specifically the subset of videos from the Perception Test that have point tracking annotations; see details in Table~\ref{tab:point_tracks}.
\begin{table}[]
    \centering
    \begin{tabular}{l|r|r}
      \textbf{Split} & \textbf{\# videos} & \textbf{\# point tracks} \\
       \hline
      Train & 28  & 1758 \\
      Validation  & 73 & 4362 \\
      Test  & 44 & 2527 \\
      \hline 
    \end{tabular}
    \caption{Dataset used for the point tracking task.}
    \label{tab:point_tracks}
\end{table}

\noindent \textbf{Baselines:} We provide baseline results for this task using a dummy static baseline, which always assumes that the point is static and visible in all frames.

\begin{figure}
    \centering
    \includegraphics[width=\linewidth]{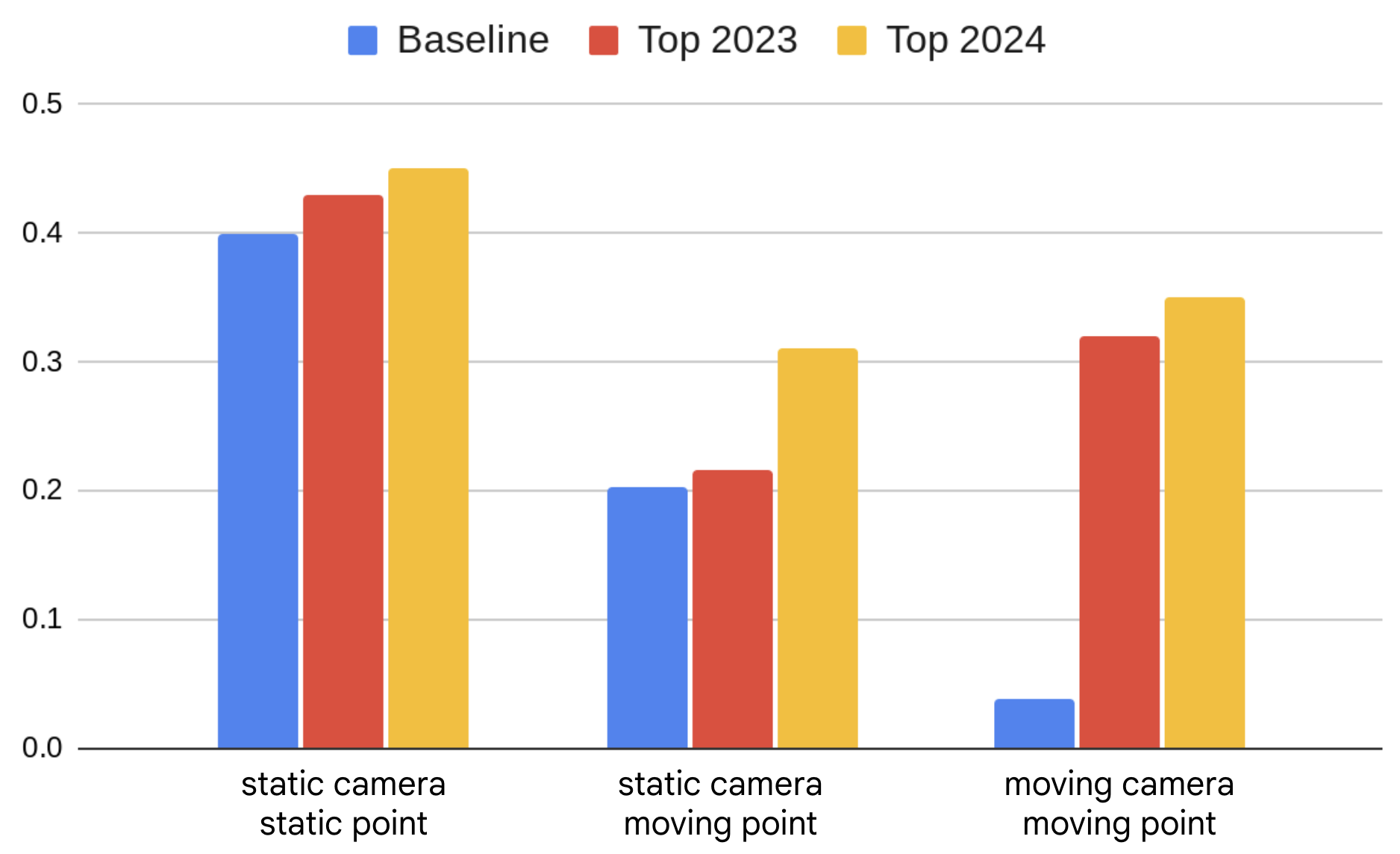}
    \caption{Baseline vs best results 2023 vs best results 2024 split by camera and point motion for the point tracking task.}
    \label{fig:pts}
\end{figure}

\noindent \textbf{Results:}
Table~\ref{tab:point_tracking} shows the results of the top-2 competing models compared to our static dummy baseline. The best results were obtained by SV (v0.6) using the LocoTrack model~\citep{cho2024localallpaircorrespondencepoint} that performs tracking of all points simultaneously, leveraging bidirectional correspondence and matching smoothness constraints -- these bring significant improvement especially for the case where the camera is static and the points are moving; see Figure~\ref{fig:pts}. 
Please check the workshop website for more details on the method included in the submission report.

\begin{table}[]
    \centering
    \begin{tabular}{l|l|c}
       \textbf{Rank} & \textbf{Team name} & \textbf{Jaccard} \\
       \hline
       Baseline & Dummy static & 0.418 \\
       Runner-up & NJUST\_kmg & 0.472 \\
       Best & SV (v0.6) & 0.474 \\
         \hline
    \end{tabular}
    \caption{Point tracking results}
    \label{tab:point_tracking}
\end{table}

\subsection{Temporal action localisation}

\noindent \textbf{Task description:} In the temporal action localisation task, the model receives a video and is required to localise and classify the actions occurring in the video according to a predefined set of classes; there are 63 action classes in total.

\noindent \textbf{Metric:} The evaluation metric for this challenge is mean average precision (mAP). It is calculated as the average precision over different action classes and IoU thresholds. For the IoU thresholds in evaluation we use [0.1 $\rightarrow$ 0.5] with 0.1 increments, similar to~\citep{Damen2021TheED}.

\noindent \textbf{Dataset:} We use the videos from the Perception Test for this challenge, as in the 2023 edition. To facilitate experimentation, we also provide features for the video / audio modalities that participants could optionally use for their submissions: video features extracted using TSP~\citep{alwassel2021tsp} and audio features extracted using MMV~\citep{alayrac2020self}. 

\noindent \textbf{Baselines:} The baseline for this task is ActionFormer~\citep{zhang2022actionformer} that we fine-tuned for the set of classes present in our benchmark.

\begin{figure*}
    \centering
    \includegraphics[width=1.\linewidth]{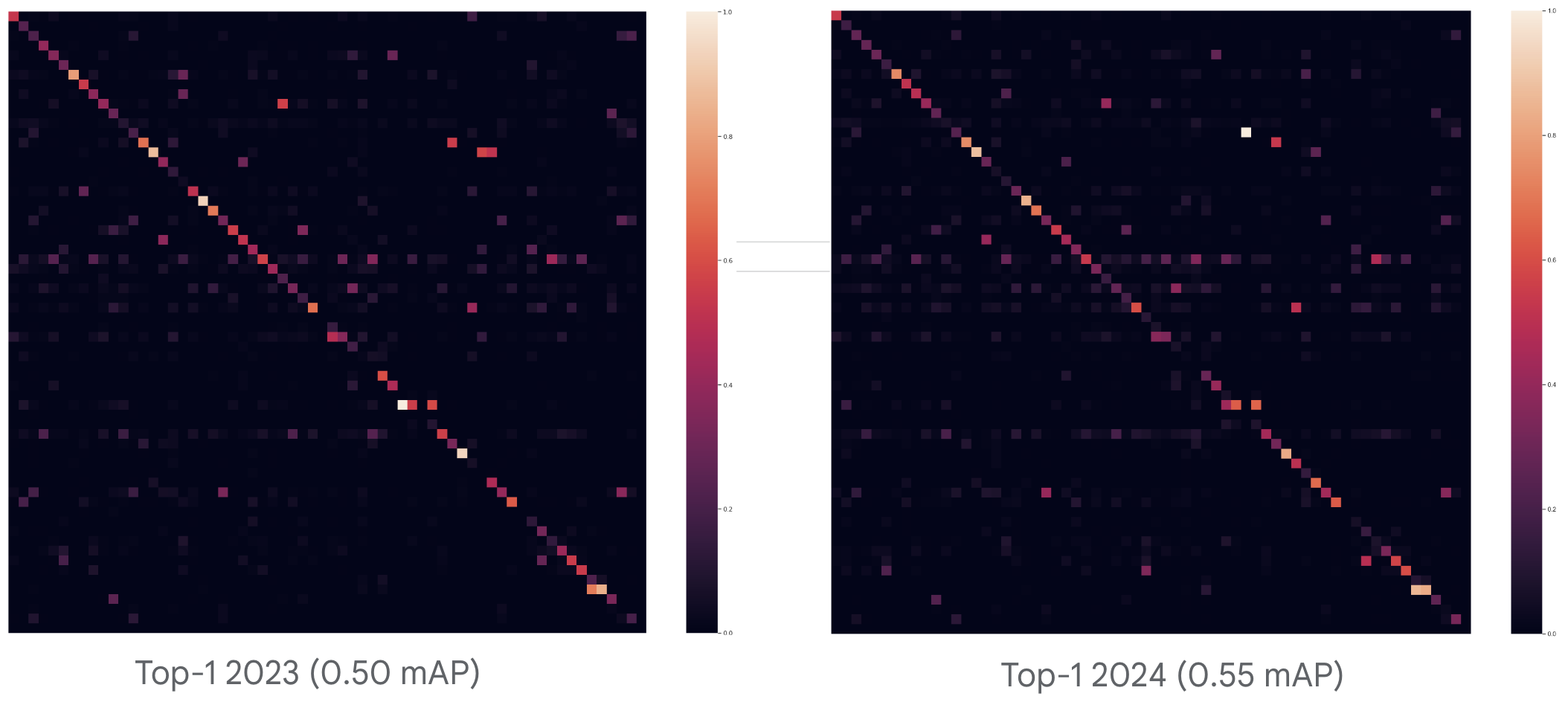}
    \caption{Confusion matrix of the best 2023 submission (left) vs best 2024 submission (right) for the temporal action localisation task. To be considered as a prediction for a certain segment, the model's confidence has to be above 0.1 and IoU threshold between the prediction and ground truth above 0.1. Ground truth actions are listed on the y-axis, sorted by their frequency and entries are normalised by rows.}
    \label{fig:tal}
\end{figure*}

\noindent \textbf{Results:} 
The results of the top-2 competing methods are included in Table~\ref{tab:tal} and are compared against our baseline. Figure~\ref{fig:tal} shows the confusion matrices of the best 2024 submission and best 2023 submission. 

The top entry this year was submitted by NJUST--\_KMG Team and uses a multimodal ActionFormer with video features obtained from UMT~\citep{liu2022umt} and VideoMAEv2~\citep{wang2023videomaev2} and audio features from BEATS~\citep{beats} and CAV-MAE~\citep{gong2023contrastive}. Please check the authors' report on our workshop page for more details.   

\begin{table}[]
    \centering
    \begin{tabular}{l|l|c}
       \textbf{Rank} & \textbf{Team name} & \textbf{mAP} \\
       \hline
       Baseline & ActionFormer & 0.156\\
       Runner-up & AITC (test\_wbf\_mamba) & 0.518 \\
       Best & NJUST--\_KMG & 0.550\\
         \hline
    \end{tabular}
    \caption{Temporal action localisation results}
    \label{tab:tal}
\end{table}

\subsection{Temporal sound localisation}

\noindent \textbf{Task description:} In the temporal sound localisation task, the model receives a video and is required to localise and classify the sound events occurring in the video according to a predefined set of sound classes; there are 16 sound classes in our dataset. For the challenge, we consider only 12 classes, excluding classes like \textit{Background}, \textit{Background-Other}, \textit{Human-Other}, \textit{Animal-Other} due to their ambiguity. 

\noindent \textbf{Metric:} Similar to the action localisation task above, the metric for this challenge is mean average precision (mAP). It is calculated as the average precision over different sound classes and IoU thresholds. For the IoU thresholds in evaluation we use [0.1 $\rightarrow$ 0.5] with 0.1 increments.

\noindent \textbf{Dataset:} As for the temporal action localisation task above, we provide the same features for all the videos in the Perception Test. 

\noindent \textbf{Baselines:} We provide baseline results for this task using the same model as in the action localisation task ActionFormer~\citep{zhang2022actionformer}, adapted to the sound localisation task by fine-tuning on our sound annotations belonging to the train split.

\begin{figure}
    \centering
    \includegraphics[width=1.\linewidth]{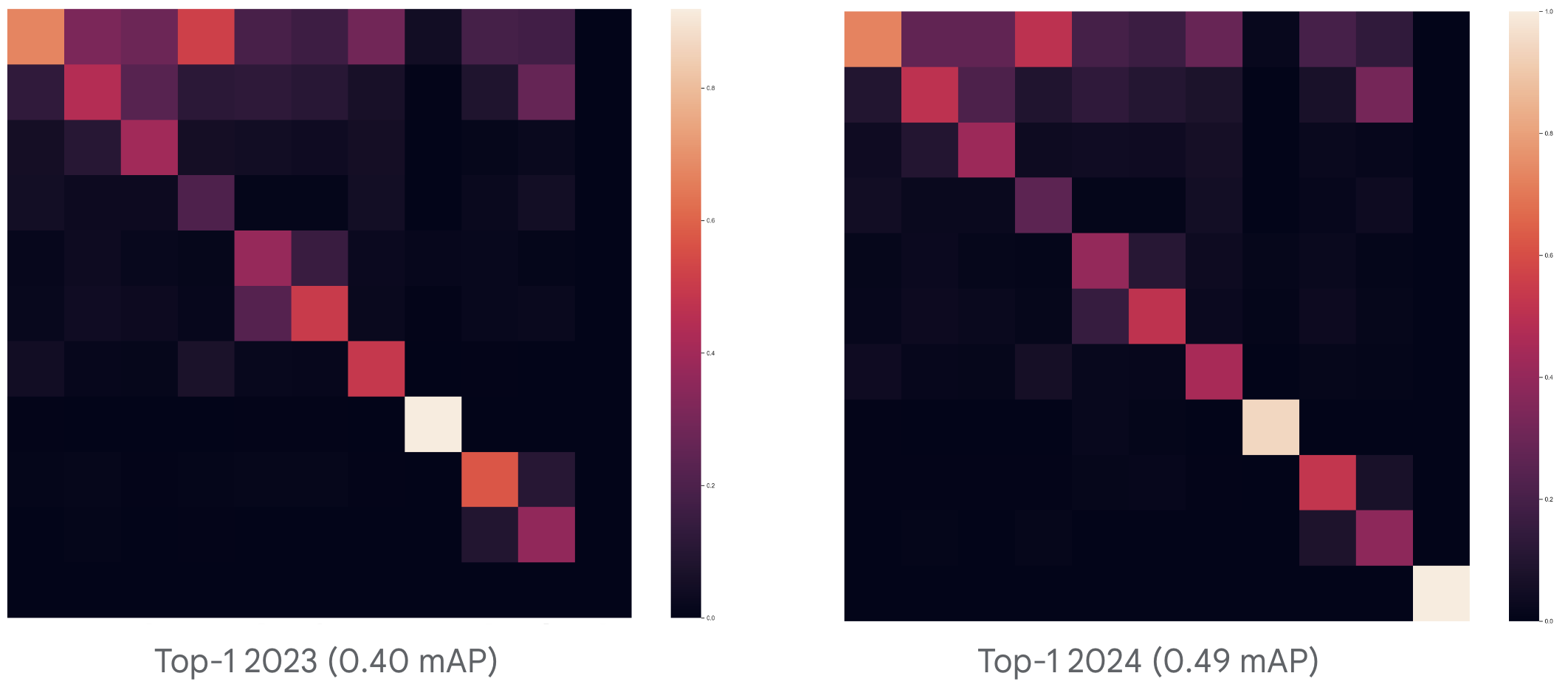}
    \caption{Confusion matrices of the best 2023 submission (left) vs best 2024 submission (right) for the temporal sound localisation task. The ground truth classes are listed on the y-axis, ordered by frequency, with scores being normalized over rows.}
    \label{fig:tsl}
\end{figure}

\noindent \textbf{Results:}
Table~\ref{tab:tsl} shows the performance of the top-2 competing methods in this track, compared to our baseline (ActionFormer). Figure~\ref{fig:tsl} compares the confusion matrices of the best model in 2024 and best submission in 2023. The 2024 top entry was submitted by NJUST\_KMG0 team and relies on an ActionFormer architecture with video features extracted using VideoMAE~\citep{videoMAE} and UMT-Large~\citep{li2023unmasked}, and audio features using BEATS~\citep{beats} and two variants of CAV-MAE~\citep{gong2023contrastive} fine-tuned on AudioSet and VGGSound, respectively. The video and audio features from all these models are extracted independently and concatenated to form the input for ActionFormer, with the audio modality having a larger number of features compared to the video, which the authors found to enhance performance; check the workshop website for more details. 

\begin{table}[]
    \centering
    \begin{tabular}{l|l|c}
       \textbf{Rank} & \textbf{Team name} & \textbf{mAP} \\
       \hline
       Baseline & ActionFormer & 0.102 \\
       Runner-up & JNU-Boat & 0.461 \\
       Best & NJUST\_KMG0 & 0.493 \\
         \hline
    \end{tabular}
    \caption{Temporal sound localisation results.}
    \label{tab:tsl}
\end{table}

\subsection{Multiple-choice video QA}

\noindent \textbf{Task description:} In the multiple-choice video question-answering (mc-vQA) task, the model receives, in parallel with the video, a question and three possible answers, out of which only one is correct, and the model has to pick one answer. The questions cover four skill areas (Memory, Abstraction, Physics, Semantics) and require different types of reasoning (Descriptive, Explanatory, Predictive, Counterfactual), across video, audio, and text modalities. The questions are also tagged with skills in each area such as: event recall (Memory), object counting (Abstraction), collision (Physics), action recognition (Semantics) and more.

\noindent \textbf{Metric:} The evaluation metric for this challenge is top-1 accuracy. It is calculated as the percentage of questions where the model's predicted option id (1 out of 3) matches the ground truth option id.

\noindent \textbf{Dataset:}
We use the same set of videos and questions as in the 2023 challenge. Recall that each video in the dataset has a number of multiple-choice video QA tasks associated, each question having 3 options, out of which only one is correct. 

\noindent \textbf{Baselines:} We provide baseline results for this task using a dummy frequency-based baseline, with multiple setups: 0-shot, few-shot, all-shot.

\begin{figure*}[t]
    \centering
    \includegraphics[width=.48\linewidth]{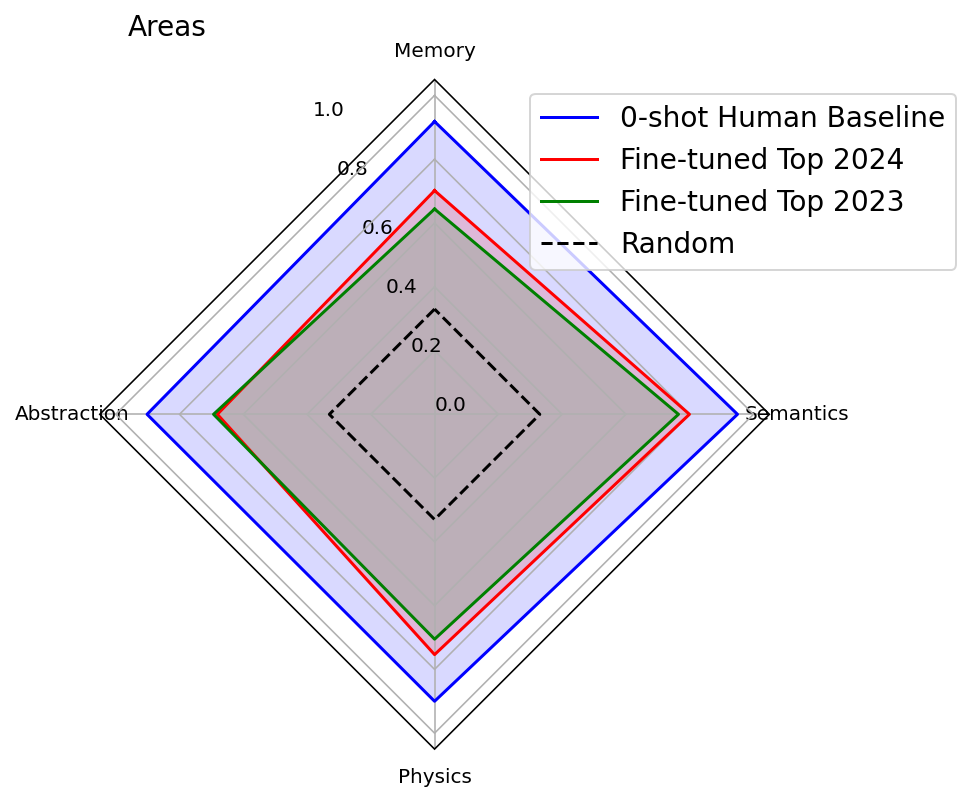}
    \includegraphics[width=.48\linewidth]{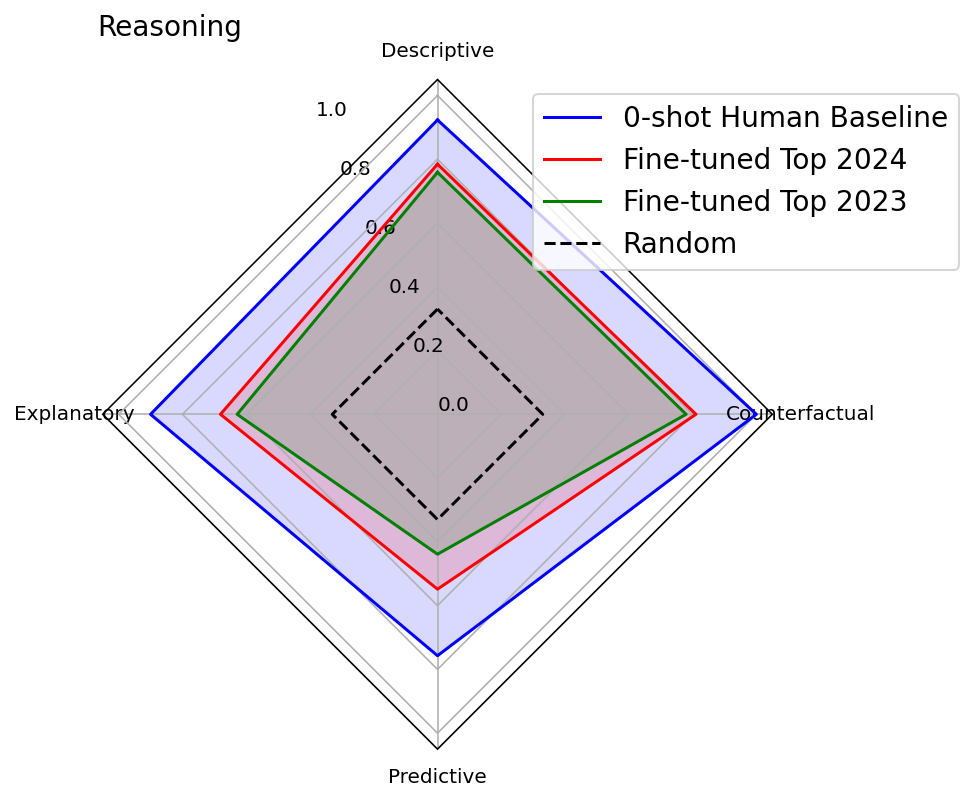}
    \caption{Random and human baselines vs best 2023 vs best 2024 detailed by areas and types of reasoning for the multiple-choice video QA task.}
    \label{fig:mcqa}
\end{figure*}

\begin{figure*}[t]
    \centering
    \includegraphics[width=\linewidth]{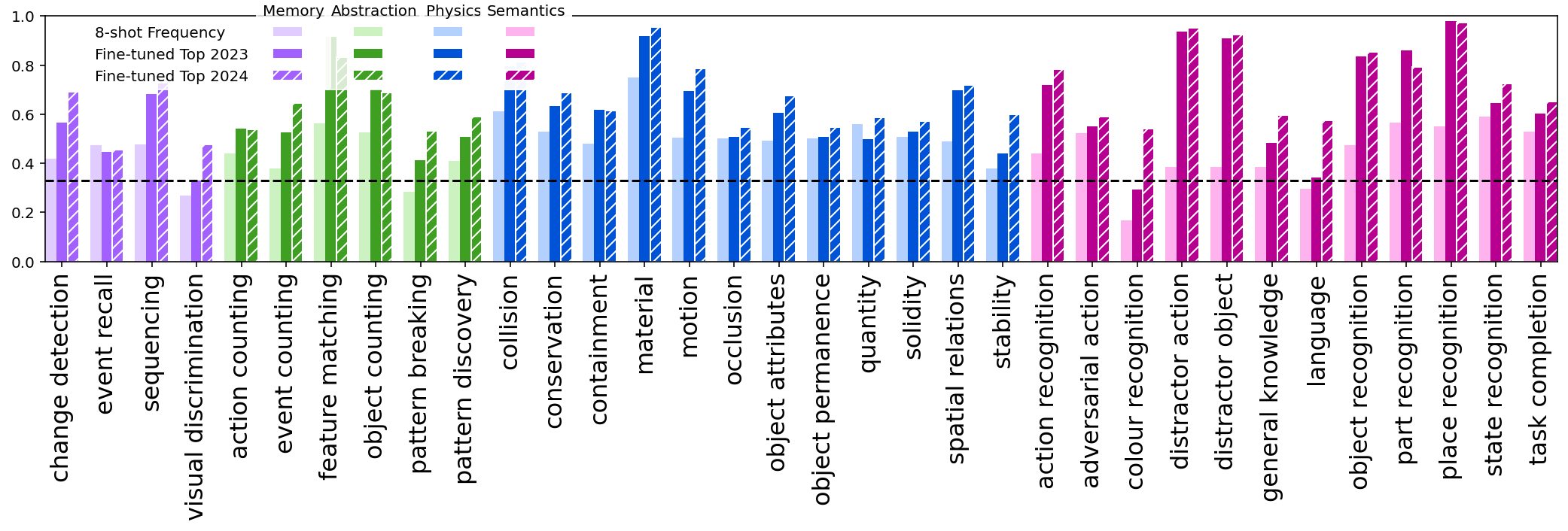}
    \caption{Baseline vs best model 2023 vs best model 2024 detailed by skills for the multiple-choice video QA task.}
    \label{fig:mcqa2}
\end{figure*}

\noindent \textbf{Results:}
Table~\ref{tab:mcqa} shows the performance of the top-2 competing models compared to our frequency baselines.

Both top-2 competing models relied on the same model, namely QwenVL2 (7B)~\citep{Qwen2VL} fine-tuned on our provided training set. The best performing model employed test-time augmentation and ensembling, whilst the runner-up used hard mining and options shuffling during fine-tuning.

Figure~\ref{fig:mcqa} shows the performance of the best 2024 submission compared to the top 2023 submission. We can observe small improvements in Physics, Memory, and Semantics, with more noticeable improvement in the Predictive reasoning type. When detailed per skill (Figure~\ref{fig:mcqa2}), we see small improvements across almost all skills.

However, Figure~\ref{fig:mcqa} shows that there is still a significant gap compared to the human baseline, which, importantly, is collected in a zero-shot setting, i.e.\ the human participants received no specific training to perform the task as detailed in the original Perception Test paper~\citep{patraucean2023perception}.  

\begin{table}[]
    \centering
    \begin{tabular}{l|l|c}
       \textbf{Rank} & \textbf{Team name} & \textbf{top-1} \\
       \hline
       Baseline 1 & Frequency (0-shot) & 0.335 \\
       Baseline 2 & Frequency (8-shot) & 0.510 \\
       Baseline 3 & Frequency (all-shot) & 0.552 \\
       Runner-up & TTgogogo (fine-tuned) & 0.764 \\
       Best & SEU-2023 (fine-tuned) & 0.765 \\
         \hline
    \end{tabular}
    \caption{Multiple-choice video QA results.}
    \label{tab:mcqa}
\end{table}

\subsection{Grounded video QA}

\noindent \textbf{Task description:} In the grounded video QA task, the model receives a video and a question/query as input, and it is required to track throughout the video the object(s) that represent the answer to the question. This is a novel type of grounded video QA task.

\noindent \textbf{Metric:} The evaluation metric for this track is HOTA (Higher Order Tracking Accuracy)~\citep{luiten2020IJCV}. It unifies the detection, association, and localization accuracy into a single metric.

\noindent \textbf{Dataset:} We use the videos from the Perception Test that have annotations for this task matching the 2023 dataset.

\noindent \textbf{Baselines:} We provide a simple baseline that runs MDETR detector~\cite{kamath2021mdetr} on the middle frame of the video using the given question as query, then it keeps the detections static throughout the video.

\begin{figure}
    \centering
    \includegraphics[width=\linewidth]{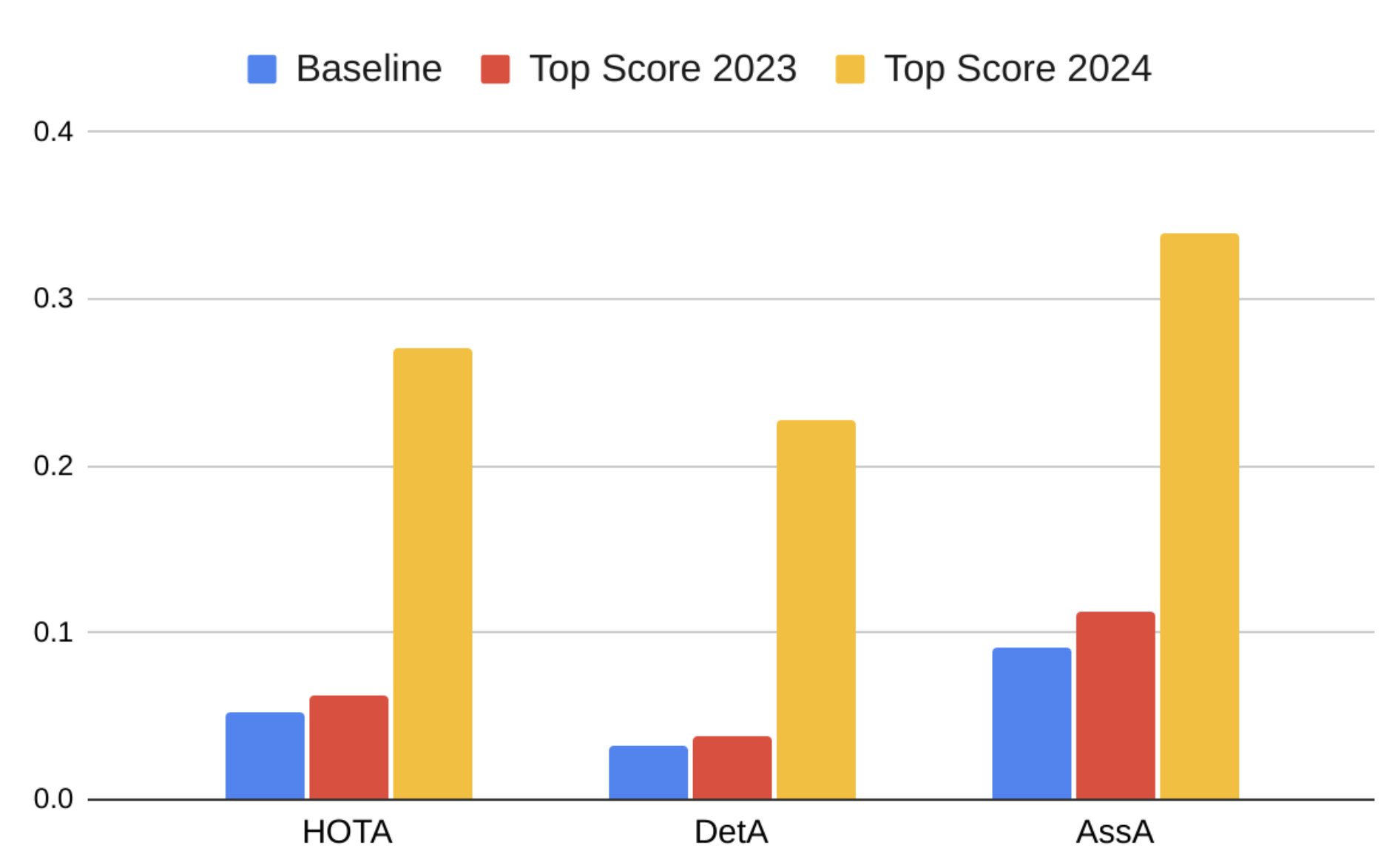}
    \caption{Baseline vs best results 2023 vs best results 2024 in terms of overall HOTA, detection, and assignment accuracy for the grounded video QA task.}
    \label{fig:hota}
\end{figure}

\noindent \textbf{Results:}
The top-2 results for this track are included in Table~\ref{tab:gqa} compared to our baseline. The top model used Gemini for obtaining a language answer to the provided question, which was then grounded using Grounding DINO~\citep{liu2023grounding}; finally, the predictions were tracked over time using SAM2~\citep{ravi2024sam2}. The runner-up solution used a similar combination of 3 components, with Llava-OneVision~\citep{li2024llava} in charge of question-answering, OWLv2~\citep{minderer2023scaling} for grounding the answers, and SAM2~\citep{ravi2024sam2} for tracking. Figure~\ref{fig:hota} compares the top model to the best 2023 submission, showing a significant improvement in performance.  

\begin{table}[]
    \centering
    \begin{tabular}{l|l|c}
       \textbf{Rank} & \textbf{Team name} & \textbf{HOTA} \\
       \hline
       Baseline & MDETR+static & 0.057 \\
              Runner-up & UCF\_CRCV & 0.241 \\
       Best & Research newbie & 0.270 \\
         \hline
    \end{tabular}
    \caption{Grounded video question-answering results.}
    \label{tab:gqa}
\end{table}

\subsection{Hour-long video QA}

\noindent \textbf{Task description:} 

In the hour-long video question-answering task, the model receives, in parallel with the video, a question and five possible answers, out of which only one is correct, and the model has to pick one answer.

\noindent \textbf{Metric:} The evaluation metric for this challenge is top-1 accuracy. It is calculated as the percentage of questions where the model's predicted option id (1 out of 5) matches the ground truth option id.

\noindent \textbf{Dataset:} We use the \textit{\wt}\ benchmark introduced in section~\ref{sec:wt}.

\noindent \textbf{Baselines:} We consider the dummy random baseline for this task, which obtains 20\%. We also provide a zero-shot human baseline: each question in the dataset was answered by 10 participants. Each participant received 27 questions. The average time for completing the batch of 27 questions was 3h50m and the overall accuracy was 99.64\%.

\noindent \textbf{Results:}
The top-2 results for this track are included in Table~\ref{tab:hvqawin}, compared to the above baselines. The top submission employs Gemini together with a zero-shot chain-of-thought approach. The model extracts keywords and task clues from the questions, and processes video segments of up to 30 minutes long in a sliding-window fashion, using previous windows as context when processing the next window; please check the workshop website for more details. These results are very promising, given how challenging these questions are. However, there is still a considerable gap between the top submissions and human performance.

\begin{table}[]
    \centering
    \begin{tabular}{l|l|c}
       \textbf{Rank} & \textbf{Team name} & \textbf{HOTA} \\
       \hline
       Baseline & Random & 0.2000 \\
       Baseline & Human & \textbf{0.9964} \\
              Runner-up & JJ\_James & 0.3729 \\
       Best & blackmonkey & 0.4407 \\
         \hline
    \end{tabular}
    \caption{Hour-long video question-answering results.}
    \label{tab:hvqawin}
\end{table}

\section{Discussion}

The Second Perception Test challenge was very successful, attracting a large number of submissions from more than hundred teams across all tracks. We observe a great improvement in performance on all tracks compared to last year, especially in the grounded video QA track where the 2023 best submission struggled to outperform a basic baseline. In addition, the newly-added track on hour-long video QA received strong submissions, showing promising hour-long video understanding capabilities. The proposed small-scale benchmark \textit{\wt}\ was created through a manual annotation collection process, but we hope that it can inspire the creation of larger-scale hour-long challenging benchmarks by, e.g., running first specialised event detectors and then designing questions based on these detections. For next year's edition of the challenge, we plan to further emphasise the zero-shot evaluation regime and incentivise participants to use a single model for addressing all tracks -- in the spirit of the original Perception Test.

\subsection*{Acknowledgements} We would like to thank Relja Arandjelovic for reviewing this report. We are grateful to Google DeepMind for providing the funding for the awards and to Ashwani Sharma from Google.org and Elder Bromley from AimGroup for ensuring a smooth handling of the awards. Special thanks to the Eval AI team for their support while running the challenges.

\bibliography{main}

\appendix

\section{Appendix}

\begin{table*}[]
    \centering
    \tiny{
    \begin{tabular}{r|p{15cm}}
         \hline
         1 & How many statue figures were there above the gate seen just before the DIESEL fashion store? \\
2 & The person holding the camera walks around a block structure with drawings on it. What did the drawings contain in the order in which they were seen?   \\
3 & The person holding the camera goes around a block structure with drawings on it. While walking around this block, a young woman is seen lighting a cigar. What kind of drawing is on the side facing this woman? \\
4 & At some point during the video, the time can be inferred from a bell ringing. How many times did the bell ring and what was the time of day? \\
5 & When passing by Mulligans pub, there is a couple coming from the opposite direction on the same side as the pub and the woman is wearing a green outfit. Where was this couple first seen and what were they doing then? \\
6 & Which of the following is true about the moment when the person holding the camera enters the Mulligan's pub? \\
7 & In which of these time intervals does the person holding the camera walk down the stairs? \\
8 & When passing by HSBC bank, there are two men walking, one of them carrying a guitar case and some other rectangular case. Where did you see them before and what was different about them then? \\
9 & In this video, when does the person holding the camera walk up the stairs? \\
10 & In which of these time intervals does the person holding the camera walk up the stairs? \\
11 & When the person enters Mohamed Ali Lane, what murals appear in order on the right wall as the person walks down the street? \\
12 & How many potted plants appear in front of the terrace next to Landesmuseum on each side? \\
13 & The Fraumunster Church clocktower appears twice. Which time is it viewed from across the water and what time of day did it show then? \\
14 & How many dogs did the person encounter during the video? \\
15 & The person holding the camera passes by two outdoor places where people sell goods, the first one around 33:20 timestamp and the second one around 51:00 timestamp. Which place was more crowded? \\
16 & The person holding the camera crosses Ponte della Paglia twice taking glimpses of Bridge of Sighs. How much time passed between the 2 crossings? \\
17 & Which mural appears two times on the person's walk around Mohamed Ali Lane? \\
18 & What types of cats are seen sitting atop of cars at less than 3 minutes distance from each other in order? \\
19 & How many moving trams appear in the first 6 minutes of the video? \\
20 & How many times does the person walk past an H\&M? \\
21 & The person crosses the street at the crossing with McDonalds around 47:00 timestamp. At what value did the traffic light counter start counting downwards? \\
22 & Around 32:40 timestamp, the person holding the camera enters in a hotel. Which of the following statements is correct? \\
23 & The person holding the camera crosses multiple bridges, sometimes crossing more than once the same bridge. The bridge crossed around 27:00 timestamp is it the same as the one crossed around 7:00 timestamp?  \\
24 & The person enters two churches in the first half hour of the video. Which one has stained glass windows and which one has clear glass windows? \\
25 & The person holding the camera passes by a group of dancers in a large plaza. Which of the following statements is correct? \\
26 & The person holding the camera films a group of musicians: 2 dressed in white shirts, one in striped shirt and one in black tshirt. In which order do these happen in the background? \\
27 & The person holding the camera films a group of four musicians. What were they wearing? \\ 
28 & In which order were the following landmarks visited? \\
29 & Around what time of the video is there an ambulance heard? \\
30 & Around what time of the day is there an ambulance heard? \\
31 & Which statement is correct? \\
32 & What time of day does the second clock-tower filmed by the person indicate? \\
33 & What times of day do the first two clock-towers encountered indicate? \\
34 & Around what time of day did the tour start? \\
35 & Which of the following statements is true? \\
36 & Which of the following statements is false? \\
37 & A mural depicting a pink pig is filmed by the person holding the camera. Which of the following statements is correct? \\
38 & How many women could be seen boarding the tram when the person holding the camera was crossing a bridge for the first time?  \\
39 & After passing by Zorba restaurant and turning on the street to the left, how many uber delivery people did the person holding the camera encounter on that street? \\
40 & What does the person holding the camera do after crossing the bridge around 28:00 timestamp? \\
41 & When passing by Caffé Nero, what sound can be heard? \\
42 & How many solo guitar buskers appear throughout the video and where? \\
43 & There is a busker with a guitar in the outdoor mall wearing a cup. What song are they singing? \\
44 & How many other fountains did the person cross by before the fountain in front of the Royale Chulan Hotel? \\
45 & How many fountains did the person cross by in total in the video? \\
46 & Which of the following is true about the first temple visited where chanting can be heard? \\
47 & The person crosses a wooden bridge twice in the video. Which time are there more people on the bridge? \\
48 & What can be seen when the camera looks to the left while crossing the bridge before reaching the quartet playing in front of Paolo Sarpi statue? \\
49 & What statue is behind the quartet playing music? \\
50 & What can be seen when the camera looks to the left while crossing the first bridge after passing by the quartet playing in front of Paolo Sarpi statue? \\
51 & At what time of day does the tour start? \\
52 & How many flags are on the fence in front of Defence Energy Department? \\
53 & A person with mime face paint is seen buttoning up their shirt by a canal. Where is this person seen again? \\
54 & The first time the person entered in a church, how much time did they spend inside? \\
55 & Of all the times times when the person sees a group of swans on the water, which time are there more adult swans? \\
         \hline
    \end{tabular}}
    \caption{List of unique questions in the proposed hour-long video QA benchmark using Walking Tours videos. Some questions were used over multiple videos resulting in the total of 70 QAs.}
    \label{tab:questions}
\end{table*}

\end{document}